\documentclass{article}
\usepackage[utf8]{inputenc}
\usepackage{spconf}
\ninept
\usepackage{myStyle}

\title{Low-rank on Graphs plus Temporally Smooth Sparse Decomposition for Anomaly Detection in Spatiotemporal Data}
\name{Seyyid Emre Sofuoglu and Selin Aviyente \thanks{This work was in part supported by NSF CCF-1615489 and DMS-1924724.}}
\address{ Department of Electrical and Computer Engineering, Michigan State University, East Lansing, MI 48823.\\sofuoglu@msu.edu, aviyente@egr.msu.edu}
 
\begin{document}

\maketitle

\begin{abstract}
Anomaly detection in spatiotemporal data is a challenging problem encountered in a variety of applications including hyperspectral imaging, video surveillance, and urban traffic monitoring. Existing anomaly detection methods are most suited for point anomalies in sequence data and cannot deal with temporal and spatial dependencies that arise in spatiotemporal data. In recent years, tensor-based methods have been proposed for anomaly detection to address this problem. These methods rely on conventional tensor decomposition models, not taking the structure of the anomalies into account, and are supervised or semi-supervised. We introduce an unsupervised tensor-based anomaly detection method that takes the sparse and temporally continuous nature of anomalies into account. In particular, the anomaly detection problem is formulated as a robust low-rank + sparse tensor decomposition with a regularization term that minimizes the temporal variation of the sparse part, so that the extracted anomalies are temporally persistent. We also approximate rank minimization with graph total variation minimization to reduce the complexity of the optimization algorithm.  The resulting optimization problem is convex, scalable, and is shown to be robust against missing data and noise. The proposed framework is evaluated on both synthetic and real spatiotemporal urban traffic data and compared with baseline methods.
\end{abstract}

\section{Introduction}
\vspace{-.5em}
Large volumes of spatiotemporal data are ubiquitous  in a diverse range of applications including climate science, social sciences, neuroscience, epidemiology \cite{bhunia2013spatial}, and transportation systems \cite{djenouri2019survey}.  Detecting anomalies from these large data volumes is important for identifying interesting but rare phenomena, e.g. traffic congestion or irregular crowd movement in urban areas. Traditional anomaly detection has focused on detecting point anomalies from time sequence data   \cite{chandola2009anomaly,fanaee2016tensor,hodge2004survey}. These methods are often  not effective for spatiotemporal data as anomalies can no longer be modeled as  i.i.d. 

In this paper, we represent spatiotemporal  data using higher-order tensors with modes corresponding to time, location and multivariate features. In this manner, we can capture both within and between-mode correlations. In particular, we focus on extracting anomalies that have spatial sparsity, \textit{i.e.}, the local changes are sparse in the spatial domain, and temporal persistence, \textit{i.e.}, the local changes last for a reasonably long time period. In recent work, we proposed Low-rank plus Temporally Smooth Sparse Decomposition (LOSS)  \cite{sofuoglu2020gloss} to incorporate these two objectives into robust low-rank + sparse tensor decomposition. First, we assumed that anomalies lie in the sparse part of the tensor, $\S$, based on the spatial sparsity property. Next, we added a regularization term that controls the temporal continuity of $\S$ to ensure temporal persistence.  In the current paper, we extend LOSS in some key ways. First, inspired by low-rank matrix recovery on graphs \cite{shahid2016fast}, we approximate low-rank tensor recovery, \textit{i.e.} nuclear norm minimization, with a graph total variation minimization across each mode. This reformulation results in a fast and scalable algorithm. Second, exploiting the manifold
information in the form of a graph incorporates
local proximity information of the data samples
into the dimensionality reduction framework, that can enhance
the anomaly detection performance in the low-dimensional space.

\noindent \textbf{Relationship to Existing Work:}
Existing tensor based anomaly detection methods have multiple shortcomings. First, they \cite{fanaee2016tensor} are mostly supervised or semi-supervised relying on historical data.  Unsupervised tensor-based anomaly detection methods \cite{xu2019anomaly}, on the other hand, aim to learn spatiotemporal features within a representation learning framework \cite{fanaee2016event,xu2019anomaly,shi2015stensr}. The learned features, \textit{i.e.} factor matrices or core tensors, are then used to detect anomalies by  monitoring the reconstruction error at each time point \cite{papalexakis2012network,papalexakis2014spotting,sun2006beyond,xu2019anomaly} or by applying well-known statistical tests to the extracted multivariate features \cite{fanaee2016event,zhang2016tensor}. Second, current methods rely on well-known low-rank tensor approximation models such as Tucker \cite{fanaee2016event,zhang2016tensor,xu2019anomaly} CP \cite{li2019tensor}, higher order RPCA (HoRPCA) \cite{li2015low,geng2014high}, and do not explicitly consider the particular structure of anomalies.   Finally, most of the existing tensor based anomaly detection methods \cite{fanaee2016tensor} focus on projection to linear subspaces. Thus, they cannot capture the nonlinear structure of anomalies which may be better represented in smooth low-dimensional manifolds. 

The proposed method is also closely related to recent work on fast robust PCA on graphs (FRPCAG) \cite{shahid2016fast,shahid2019tensor}. This work shows that it is possible to implement low-rank matrix recovery through two graph regularization terms, \textit{i.e.} smoothness across the rows and columns of the data matrix, avoiding computationally expensive nuclear norm minimization as long as the data matrix is low-rank on graphs. In this paper, we extend this approximation to tensors and implement low-rank tensor recovery  through graph regularization across each mode of the tensor.
\vspace{-1em}
\section{Background}
\label{sec:back}
\vspace{-1em}
\subsection{Tensor Notation}
\vspace{-.5em}
Let $\Y\in\R^{I_1\times I_2\times \dots \times I_N}$ be a tensor of order $N$, where $\Y_{i_1,i_2,\dots,i_N}$ denotes the $({i_1,i_2,\dots,i_N})^{th}$ element of the tensor. 

\noindent \textbf{Definition 1.} (Mode-$n$ unfolding) The mode-n unfolding of a tensor $\Y$ is defined as $\Y_{(n)}\in \R^{I_n\times \prod_{n'=1,n'\neq n}^NI_{n'}}$ where the mode-n fibers of the tensor $\Y$ are the columns of $\Y_{(n)}$ and the remaining modes are organized accordingly along the rows.

\noindent \textbf{Definition 2.} (Mode-$n$ product) The mode-$n$ product of a tensor $\mathcal{A}\in \mathbb{R}^{I_1\times ... I_n\times ...\times I_N}$ and a matrix ${\bf{U}} \in \mathbb{R}^{J\times I_n} $ is denoted as $\mathcal{B}=\mathcal{A} \times_{n} {\bf{U}}$ and is equivalently rewritten as $\mathcal{B}_{(n)}=\mathbf{U}\mathcal{A}_{(n)}$, where $\mathcal{B} \in \mathbb{R}^{I_1\times ...\times I_{n-1} \times J \times I_{n+1}\times ...\times I_N}$. 


\noindent \textbf{Definition 3.} (Mode-$n$ Graph Laplacian) Let $W^n$ be the similarity matrix across mode-$n$, constructed as the k-nearest neighbor graph of $\Y_{(n)}$ using  a Gaussian kernel. The mode-$n$ graph Laplacian $\Phi^n$ is then defined as $\Phi^n=D^n-W^n$, where $D^n$ is a diagonal degree matrix with $D_{i,i}^n = \sum_{i'=1}^{I_n} W_{i,i'}^n$. The eigendecompositon of $\Phi^{n}$ can be written as $\Phi^n = P_n \Lambda_n P_n^\top$, where $P_n$ is the matrix of eigenvectors and $ \Lambda_n$ is a diagonal matrix with the eigenvalues on the diagonal, in a non-descending order. 

\noindent \textbf{Definition 4.} (Tensor norms)  Frobenius norm of a tensor is defined as $\|\Y\|_F=\sqrt{\sum_{i_1,i_2,\dots,i_N} \Y_{i_1,i_2,\dots,i_N}^2}$.  $\ell_1$ norm of a tensor is defined as $\|\Y\|_1=\sum_{i_1,i_2,\dots,i_N}|\Y_{i_1,i_2,\dots,i_N}|$. 

\noindent \textbf{Definition 5.} (Support Set) Let $\Omega$ be an index set defined for tensor $\Y$, \textit{i.e.} $\Omega \in [I_1]\times [I_2]\times\dots\times[I_N]$. A projection operator, $\P_\Omega$, is defined on this support set as:
\begin{gather}
    \P_\Omega[\Y]_{i_1,i_2,\dots,i_N} = 
    \begin{cases}
        \Y_{i_1,i_2,\dots,i_N}, & (i_1,i_2,\dots,i_N)\in\Omega\\
        0, & \text{otherwise.}
    \end{cases}
\end{gather}
Orthogonal complement of $\P_\Omega$ is denoted as $\P_{\Omega^\perp}$. 
\vspace{-1em}
\subsection{Robust PCA on Graphs}
\vspace{-.5em}
\label{subsec:RPCAG}
In many modern signal processing applications, graph-based priors have been used to  extract low-dimensional structure from high dimensional data \cite{du2015sparse,jin2015low,tao2015low}. Representation of a signal on a graph is also  motivated by
the emerging field of signal processing on graphs, based on
notions of spectral graph theory \cite{shahid2016fast,shahid2019tensor,perraudin2017stationary}. The underlying assumption is that high-dimensional data samples lie on or close to
a smooth low-dimensional manifold, represented by a graph $G$. 

In \cite{shahid2016fast}, it was shown that a low-rank approximation, $U$, to a data matrix, $X$, can be obtained by solving the following optimization problem:
\begin{equation}
    \min_{U} \|X-U\|_{1}+\gamma_{1}tr(U\Phi^{1}U^{T})+\gamma_{2}tr(U^{T}\Phi^{2}U),
    \label{eq:frpcag}
\end{equation}
where $\Phi^{1}$ and $\Phi^{2}$ are the graph Laplacians corresponding to graphs connecting the samples (rows) of $X$ and the features (columns) of $X$, respectively. The above formulation assumes that the data is low-rank on graphs, \textit{i.e.} lies on a smooth low-dimensional manifold. This can be quantified by  a graph stationarity measure, $
    s_r(\Gamma_n)=\frac{\|\text{diag}(\Gamma_n)\|_2}{\|\Gamma_n\|_F}    
$, where $\Gamma_n = P_n^\top C_n P_n$ with $C_n$ being the covariance matrix of each mode-$n$ unfolding, i.e., $n=1$ or $2$ in the case of data matrices \cite{perraudin2017stationary,shahid2019tensor}. 

\vspace{-1em}
\section{Methods}
\vspace{-.5em}
\label{sec:meth}
In this paper,  $\Y$ is spatiotemporal data from which we extract the anomalous entries. The first mode corresponds to temporal intervals such as hours in a day, \textit{i.e.} $I_1=24$, while the remaining modes correspond to the different spatial regions and various features such as sensors, weeks, days, years, etc. depending on the data structure. 
\vspace{-1em}
\subsection{Problem Statement}
\vspace{-.5em}
Given an observed tensor with possibly missing entries, $\P_{\Omega}[\Y]$,  our goal is to learn a low-rank + sparse representation, where the low-rank part corresponds to the normal activity and the sparse part, $\S$,  corresponds to the anomalies. As the anomalies in urban data are generally temporally persistent, \textit{i.e.} not instantaneous, we take the temporal smoothness of the sparse part into account through a total variation (TV) regularization term, $\|\S\times_1 \Delta\|_1$, where $\Delta$ is the first order discrete-time differentiation operator. In prior work \cite{sofuoglu2020gloss}, we formulated the following objective function to accommodate these assumptions:
\begin{gather}
    \min_{\L, \S} \theta \sum_{n=1}^N\|\L_{(n)}\|_*+\lambda\|\S\|_1+\gamma\|\S\times_1 \Delta\|_1,\nonumber\\ 
   s.t. \hspace{0.5cm} \P_{\Omega}[\Y] = \P_{\Omega}[\L+\S], 
    \label{eq:obLOSS}
\end{gather}
where $\L$ is the low-rank, $\S$ is the sparse tensor and $\theta$, $\lambda$ and $\gamma$ are the regularization parameters. The solution to \eqref{eq:obLOSS} will be referred as LOSS, hereafter.

As mentioned in Section \ref{subsec:RPCAG}, we will approximate the low-rank tensor, $\mathcal{L}$, through $N$ graph total variation terms corresponding to each mode similar to FRPCAG in \eqref{eq:frpcag}. To this end, the first $J_n$ eigenvectors of $\Phi^{n}$, $\hat{P}_n$ corresponding to the $J_n$ lowest eigenvalues, are used to quantify the total variation of the low-rank tensor across mode-$n$ with respect to its corresponding similarity graph. As these first $J_{n}$ eigenvectors capture the low-frequency information of the signal, they can capture the normal activity in the data. Thus, the optimization problem can be written as:
\begin{gather}
    \min_{\L, \S} \theta \sum_{n=1}^N\mathrm{tr}\left(\L_{(n)}^\top \hat{\Phi}^n\L_{(n)}\right)+\lambda\|\S\|_1+\gamma\|\S\times_1 \Delta\|_1,\nonumber\\ 
   s.t. \hspace{0.5cm} \P_{\Omega}[\Y] = \P_{\Omega}[\L+\S], 
    \label{eq:obL}
\end{gather}
where $\hat{\Phi}^n=\hat{P}_n\hat{\Lambda}_n\hat{P}_n^\top$ and $\hat{\Lambda}_n\in\R^{J_n\times J_n}$ is the leading principal submatrix of $\Lambda_n$. If we define the projections of each mode-$n$ unfolding of $\L$ to the graph eigenvectors (low frequency graph Fourier basis) as $\G^n_{(n)}=\hat{P}_n^\top\L_{(n)}$, then \eqref{eq:obL} can be rewritten as:
\begin{gather}
    \min_{\L,\{\G\}\S}\theta \sum_{n=1}^N\mathrm{tr}\left({\G_{(n)}^n}^\top { \Lambda}^n\G_{(n)}^n\right)+\lambda\|\S\|_1+\gamma\|\S\times_1 \Delta\|_1,\nonumber\\ 
    s.t.\hspace{0.5cm}
    \P_{\Omega}[\Y] = \P_{\Omega}[\L+\S],\qquad \G^n_{(n)}=\hat{P}_n^\top\L_{(n)},  
    \label{eq:oblogss1}
\end{gather}
where $\{\G\}:=\{\G^1,\dots,\G^N\}$. The solution to \eqref{eq:oblogss1} will be called LOw-rank on Graphs plus temporally Smooth Sparse Decomposition (LOGSS).
\vspace{-1em}
\subsection{Optimization}
\vspace{-.5em}
The optimization problem was solved using ADMM, as it has been utilized in solving similar convex problems \cite{goldfarb2014robust,aggarwal2016hyperspectral,sofuoglu2020gloss}. We introduce auxiliary variables $\W$ and $\Z$ to separate sparsity and temporal smoothness regularization. The problem is then rewritten as:
\begin{gather}
    \min_{\L,\{\G\},\S,\W,\Z}\theta \sum_{n=1}^N\mathrm{tr}\left({\G_{(n)}^n}^\top { \Lambda}^n\G_{(n)}^n\right)+\lambda\|\S\|_1+\gamma\|\S\times_1 \Delta\|_1,\nonumber\\ 
    \text{s.t.} \quad
    \P_{\Omega}[\Y] = \P_{\Omega}[\L+\S],\quad \G^n_{(n)}=\hat{P}_n^\top\L_{(n)},\nonumber\\ \S=\W,\quad \Z=\W\times_1\Delta.  
    \label{eq:oblogss}
\end{gather}
The corresponding augmented Lagrangian is given by:  
\begin{gather}
    \theta\sum_{n=1}^N\mathrm{tr}\left({\G_{(n)}^n}^\top {\Lambda}_n\G_{(n)}^n\right)+\lambda\|\S\|_1+\gamma\|\Z\|_{1}+\nonumber \\ \frac{\beta_1}{2}\|\P_{\Omega}[\L+\S-\Y]- \Gamma_1\|_F^2+ \frac{\beta_2}{2}\|\W\times_1 \Delta-\Z- \Gamma_2\|_F^2+\nonumber \\ \frac{\beta_3}{2}\|\S-\W- \Gamma_3\|_F^2+\frac{\beta_4}{2}\sum_{n=1}^N\|\L-\G^n\times_n \hat{P}_n- \Gamma_4^n\|_F^2,
    \label{eq:aug_lagr}
\end{gather}
where $ \Gamma_1,  \Gamma_2,  \Gamma_3,  \Gamma_4^n \in\R^{I_1\times I_2\times I_3\times I_4}$ are the Lagrange multipliers. Using \eqref{eq:aug_lagr} each variable can be updated alternately.

\noindent \textbf{1. $\L$ update:} The update of low-rank variable $\L$ is given by:
\begin{gather}
    \P_{\Omega}[\L^{t+1}] = \P_{\Omega}\left[\beta_1 \TT_1+\beta_4\TT_2\right]/(\beta_1+4\beta_4),\nonumber\\
    \P_{\Omega^\perp}[\L^{t+1}] = \P_{\Omega^\perp}[\TT_2]/4,
    \label{eq:l_up}
\end{gather}
where $\TT_1=\Y-\S^t+ \Gamma_1^t$, $\TT_2=\sum_{n=1}^N\G^{n,t}\times_n \hat{P}_n+ \Gamma_4^{n,t}$.

\noindent \textbf{2. $\G^n$ update:} The variables $\G^n$ can be updated using:
\begin{gather}
    \G^{n,t+1} = (2\frac{\theta}{\beta_4}\hat{\Lambda}_n+\I)^{-1}(\L^{t+1}\times_n \hat{P}_n^\top- \Gamma_4^{n,t}),
    \label{eq:g_up}
\end{gather}
where $\I\in \R^{J_n\times J_n}$ is an identity matrix. 

\noindent \textbf{3. $\S$ update:}
The variable $\S$ can be updated using:
\begin{gather}
    \P_{\Omega}[\S^{t+1}]=\T_{ \lambda} (\P_{\Omega}[\beta_1\TT_3+\beta_3\TT_4])/(\beta_1+\beta_3)\nonumber\\ \P_{\Omega^\perp}[\S^{t+1}]= \T_{\frac{\lambda}{\beta_3}}(\P_{\Omega^\perp}[\TT_4]),
    \label{eq:s_up}
\end{gather}
where $\TT_3\!=\!\Y\!-\!\L^{t+1}\!+\! \Gamma_1^t$, $\TT_4\! =\! \W^t\!+\! \Gamma_3^t$, $\T_{\phi}(\a) = sign(\a)\odot max(|\a|-\phi, 0)$ and $\odot$ is Hadamard product. 

\noindent \textbf{4. $\W$ update:} The auxiliary variable $\W$ can be updated using:
\begin{gather}
    \W^{t+1}_{(1)} = W_{inv}\left(\beta_3(\S- \Gamma_3)_{(1)}+\beta_2 \Delta^\top( \Gamma_2+\Z)_{(1)}\right),
    \label{eq:w_up}
\end{gather}
where $W_{inv}=\left(\beta_3\I+\beta_2 \Delta^\top \Delta\right)^{-1}$ always exists and can be computed outside the loop for faster update.

\noindent \textbf{5. $\Z$ update:} The auxiliary variable $\Z$, can be updated using:
\begin{gather}
    \Z^{t+1} = \argmin_{\Z} \gamma\|\Z\|_{1}+\frac{\beta_2}{2}\|\W^{t+1}\times_1 \Delta-\Z- \Gamma_2^t\|_F^2,
    \label{eq:z_up}
\end{gather}
which is solved by $\T_{\frac{\gamma}{\beta_2}}(\W^{t+1}\times_1 \Delta- \Gamma_2^t)$.

\noindent \textbf{6. Dual updates:} Finally, dual variables $ \Gamma_1,  \Gamma_2,  \Gamma_3,  \Gamma_4^n$ are updated using:
\begin{gather}
     \Gamma_1^{t+1} =  \Gamma_1^t-\P_\Omega[\L^{t+1}+\S^{t+1}-\Y],\label{eq:G_1}\\
     \Gamma_2^{t+1} =  \Gamma_2^{t}-(\W^{t+1}\times_1 \Delta-\Z^{t+1}),\label{eq:G_2}\\
     \Gamma_3^{t+1} =  \Gamma_3^{t}-(\S^{t+1}-\W^{t+1}),\label{eq:G_3}\\
     \Gamma_4^{n,t+1} =  \Gamma_4^{n,t}-(\L^{t+1}-\G^{n,t+1}\times_n \hat{P}_n).\label{eq:G_4}
\end{gather}
The pseudocode for the optimization is given in Algorithm \ref{alg:logss}.
\setlength{\textfloatsep}{5pt}
\renewcommand{\algorithmicrequire}{\textbf{Input:}}
\renewcommand{\algorithmicensure}{\textbf{Output:}}
\begin{algorithm}  
\caption{LOGSS}
\begin{algorithmic}
\REQUIRE $\Y$, $\Omega$, $\Phi_n$, parameters $\theta$, $\lambda$, $\gamma$, $\beta_1$, $\beta_2$, $\beta_3$, $\beta_4$, $\text{max\_iter}$.
\ENSURE $\L$ : Low-rank tensor;  $\S$: Sparse tensor.
\STATE Initialize $\S^0=0$, $\W^0=0$, $\Z^0=0$, $\G^{n,0}=0$, $ \Gamma_1^0=0$, $ \Gamma_2^{0}=0$, $ \Gamma_3^{0}=0$, $ \Gamma_4^{n,0}=0$, $\forall i \in \{1,\dots,4\}$, $W_{inv}=\left(\beta_3\I+\beta_2 \Delta^\top \Delta\right)^{-1}$.
\FOR{$t=1$ \TO $\text{max\_iter}$}
\STATE{Update $\L$ using \eqref{eq:l_up}.}
\STATE{Update $\G^n$s using \eqref{eq:g_up}.}
\STATE{Update $\S$ using \eqref{eq:s_up}.}
\STATE{Update $\W$ using \eqref{eq:w_up}.}
\STATE{Update $\Z$ using \eqref{eq:z_up}.}
\STATE{Update Lagrange multipliers using \eqref{eq:G_1}, \eqref{eq:G_2}, \eqref{eq:G_3} and \eqref{eq:G_4}.}
\ENDFOR
\end{algorithmic}
\label{alg:logss}
\end{algorithm} 
\vspace{-1em}
\subsection{Convergence}
\vspace{-.5em}
The convergence of ADMM is proven for two-block systems and three-block systems with at least one strongly convex and two convex functions in \cite{cai2014direct,li2015convergent}. It can easily be shown using Kronecker products and vectorizations that the objective function can be converted into a two-block form. Since all parts of the objective function are convex, the proposed algorithm  converges \footnote{The readers are referred to \cite{sofuoglu2020gloss} for a detailed analysis of convergence for a similar problem.}. 
\vspace{-1em}
\subsection{Computational Complexity}
\vspace{-.5em}
Assume $I_1=I_2=\dots=I_N=I$. The complexity of the proposed algorithm is dominated by matrix multiplications which are the updates of $\L$, $\W$, $\Z$ and $ \Gamma_4^n$. The updates of $\G^n$ require are multiplications since $ \hat{\Lambda}_n$s are diagonal. The computational complexity of the matrix multiplications are: $\O(NI^N)$ for the update of $\L$, $\O(I^N)$ for the updates of $\W, \Z,  \Gamma_4^n$. Since the updates of $\L$ and $ \Gamma_4^n$ can be parallelized, the complexity of the algorithm is $\O(\text{max\_iter}I^{N})$, hence, linear in the number of elements. In comparison, approximation of low-rank tensor using nuclear norm minimization results in quadratic complexity, \textit{i.e.} $\O(\text{max\_iter}I^{2N})$ \cite{sofuoglu2020gloss}. 
\vspace{-1em}
\subsection{Anomaly Scoring}
\vspace{-.5em}
Following \cite{zhang2019decomposition} and \cite{sofuoglu2020gloss}, we applied Elliptic Envelope (EE) \cite{rousseeuw1999fast} to each third-mode fiber of the extracted sparse tensor to assign an anomaly score to each tensor element.  The anomaly scores were then used in ranking the elements and selecting the top-$K$ as anomalous elements. With varying $K$ and different initializations, the mean of the area under the curve (AUC) values were reported.
\addtolength{\parskip}{-0.5mm}
\begin{table*}[]
\scriptsize 
    \vspace{-1em}
    \centering
    \caption{Mean and standard deviation of AUC values for various $c$ (Rows 1-3), $P$ (4-5), $l$ (6-7) and $m$ (8-9). On experiments of each variable, the rest of the variables are fixed at $c=2.5$, $P=0\%$, $l=7$ and $m=2.3\%$. }
    \begin{tabular}{l|l|cc|cc|cc|cc}
    \toprule
    & &\multicolumn{2}{c}{EE} & \multicolumn{2}{c}{HoRPCA}  & \multicolumn{2}{c}{LOSS} & \multicolumn{2}{c}{LOGSS}
    \\ 
    & & AUC& Time(sec)& AUC& Time(sec)& AUC& Time(sec)& AUC& Time(sec)\\
    \midrule
    1& $c=1.5$ & $0.71\!\pm\!0.004$ & $11.6\!\pm\! 0.12$ & $0.70\!\pm\! 0.004$   & $14.9\!\pm\! 0.35$ & $\mathbf{0.81\!\pm\!0.005}$& $36.0\!\pm\! 0.9$ & $\mathbf{0.80\!\pm\!0.005}$& $\mathbf{5.2\!\pm\!0.2}$  \\
    2& $c=2$ & $0.81\!\pm\!0.004$ & $11.6\!\pm\! 0.15$ & $0.81\!\pm\! 0.004$   & $14.9\!\pm\! 0.42$ & $\mathbf{0.90\!\pm\!0.004}$& $35.8\!\pm\! 0.9$ & $\mathbf{0.90\!\pm\!0.003}$& $\mathbf{5.0\!\pm\!0.2}$  \\
    3& $c=2.5$ & $0.87\!\pm\!0.002$ & $11.6\!\pm\! 0.17$  &  $0.87\!\pm\!0.003$  & $15\!\pm\!0.41$ & $\mathbf{0.94\!\pm\!0.003}$  & $35.9\!\pm\! 0.9$ & $\mathbf{0.94\!\pm\!0.002}$ & $\mathbf{5.2\!\pm\!0.2}$\\
    \midrule
    4&$P = 20\%$ & $0.81\!\pm\!0.004$  & $12.7\!\pm\! 0.25$  & $0.80\!\pm\!0.004$&  $18.9\!\pm\! 1.6$ & ${0.85\!\pm\!0.003}$ &$35.9\!\pm\! 1.17$ & $\mathbf{0.86\!\pm\!0.004}$ & $\mathbf{4.9\!\pm\!0.18}$\\ 
    5&$P = 40\%$ &$0.61\!\pm\!0.008$ & $14.2\!\pm\! 0.35$ & $0.72\!\pm\!0.006$ & $17.6\!\pm\! 0.57$ &  $\mathbf{0.73\!\pm\!0.009}$ & $37.6\!\pm\! 1.2$ & $\mathbf{0.74\!\pm\!0.007}$  & $\mathbf{5.0\!\pm\!0.29}$  \\
    \midrule
    6&$l=5$ & $0.87\!\pm\! 0.003$ &$11.8\!\pm\!0.2$ & $0.87\!\pm\! 0.003$ & $15.2\!\pm\!0.25$ & ${0.92\!\pm\!0.004}$ & $31.0\!\pm\!0.27$ & $\mathbf{0.93\!\pm\!0.001}$ & $\mathbf{4.4\!\pm\! 0.15}$ \\
    7&$l=15$ & $0.87\!\pm\!0.002$ & $11.7\!\pm\! 0.25$ & $0.87\!\pm\! 0.002$ & $30.8\!\pm\!0.27$ & $\mathbf{0.96\!\pm\!0.002}$ & $30.8\!\pm\!0.27$ & $0.95\!\pm\! 0.001$ & $\mathbf{4.4\!\pm\! 0.05}$\\
    \midrule
    8&$m=1.7\%$ & $0.87\!\pm\! 0.004$ &$11.6\!\pm\!0.16$ & $0.87\!\pm\! 0.004$ & $14.2\!\pm\!0.27$ & $\mathbf{0.94\!\pm\!0.004}$ & $32.0\!\pm\!1.17$ & $\mathbf{0.94\!\pm\!0.003}$ & $\mathbf{4.5\!\pm\! 0.41}$ \\
    9&$m=33.4\%$ & $0.86\!\pm\!0.001$ & $11.5\!\pm\! 0.07$ & $0.86\!\pm\! 0.001$ & $15.0\!\pm\!0.18$ & $\mathbf{0.92\!\pm\!0.001}$ & $31.5\!\pm\!0.71$ & $\mathbf{0.93\!\pm\! 0.001}$ & $\mathbf{4.75\!\pm\! 0.4}$\\
    \bottomrule
    \end{tabular}
    \label{tab:nyc_auc}
    \vspace{-2em}
\end{table*}

\vspace{-1em}
\section{Experiments}
\vspace{-.5em}
We compare the proposed method to LOSS and HoRPCA on real and synthetic data sets to evaluate the improvements provided by graph total variation minimization and temporal smoothness regularization, respectively. Moreover, we compare to EE applied to raw data  to evaluate the contribution of optimization based feature extraction. 

In this paper, we used NYC yellow taxi trip records\footnote{https://www1.nyc.gov/site/tlc/about/tlc-trip-record-data.page} for 2018. The data was pre-processed to create a tensor $\Y$ of size $24\times 7\times 52 \times 81$, where modes correspond to hours of a day, days of a week, weeks of a year and selected NYC taxi zones as described in \cite{sofuoglu2020gloss}. The data is suitable for  low-rank on graphs model as graph stationarity measure $s_r(\Gamma)$ for each mode is $0.83, 0.98, 0.99, 0.56$, respectively. This implies that the data is mostly low-rank on the temporal modes as there is strong correlation among the different days, hours and weeks, while it is less low-rank across space. 
 
Following \cite{sofuoglu2020gloss} and \cite{zhang2019decomposition}, we also generate synthetic data with ground truth labels to evaluate all methods under four different conditions: missing data, noise,  number of anomalies and length of anomalies. To generate synthetic data, we take the average of $\Y$ along the third mode. We then repeat the resulting tensor with $I_3=1$ such that the resulting tensor has the same size as $\Y$. We multiply each element of the tensor by a Gaussian random variable with mean $1$ and variance $0.5$ to create variation across weeks. A percentage $P$ of all first mode fibers is set to zero to simulate missing data.

We generate anomalies on randomly selected $m\%$ of the first mode fibers. For each fiber, we  set a random time interval of length $l$, which corresponds to $l$ hours in a day, as anomalous.  We multiply the average value of each randomly selected anomalous interval by a parameter $c$ and the modify the entries by adding or subtracting this value from the interval. When $c$ is low, the anomalies will be harder to detect and may be perceived as noise.

\noindent \textbf{Parameter Selection:} To build the graph Laplacians at each mode $n$, we select $\min(10, I_n-1)$ nearest neighbor graphs with Gaussian kernel as described in Section \ref{sec:back}. The rank for each mode $n$, $J_n$ is chosen such that $J_n = \inf_{i}\{i|i\in \{1,\dots,I_n-1\}, \frac{\lambda_{i,n}}{\lambda_{i+1,n}}>.9\}$, where $\lambda_{i,n}$ is the $i$th eigenvalue of mode-$n$ graph Laplacian $\Phi^n$. All other parameters for all methods are tuned empirically for the best results. After tuning, the parameters were fixed for experiments of $c$, $P$ variables and $(l, m)$ tuples, for the results to be comparable.
\vspace{-1em}
\subsection{Experiments on Synthetic Data}
\begin{figure}
    \centering
    \includegraphics[width = .95\columnwidth]{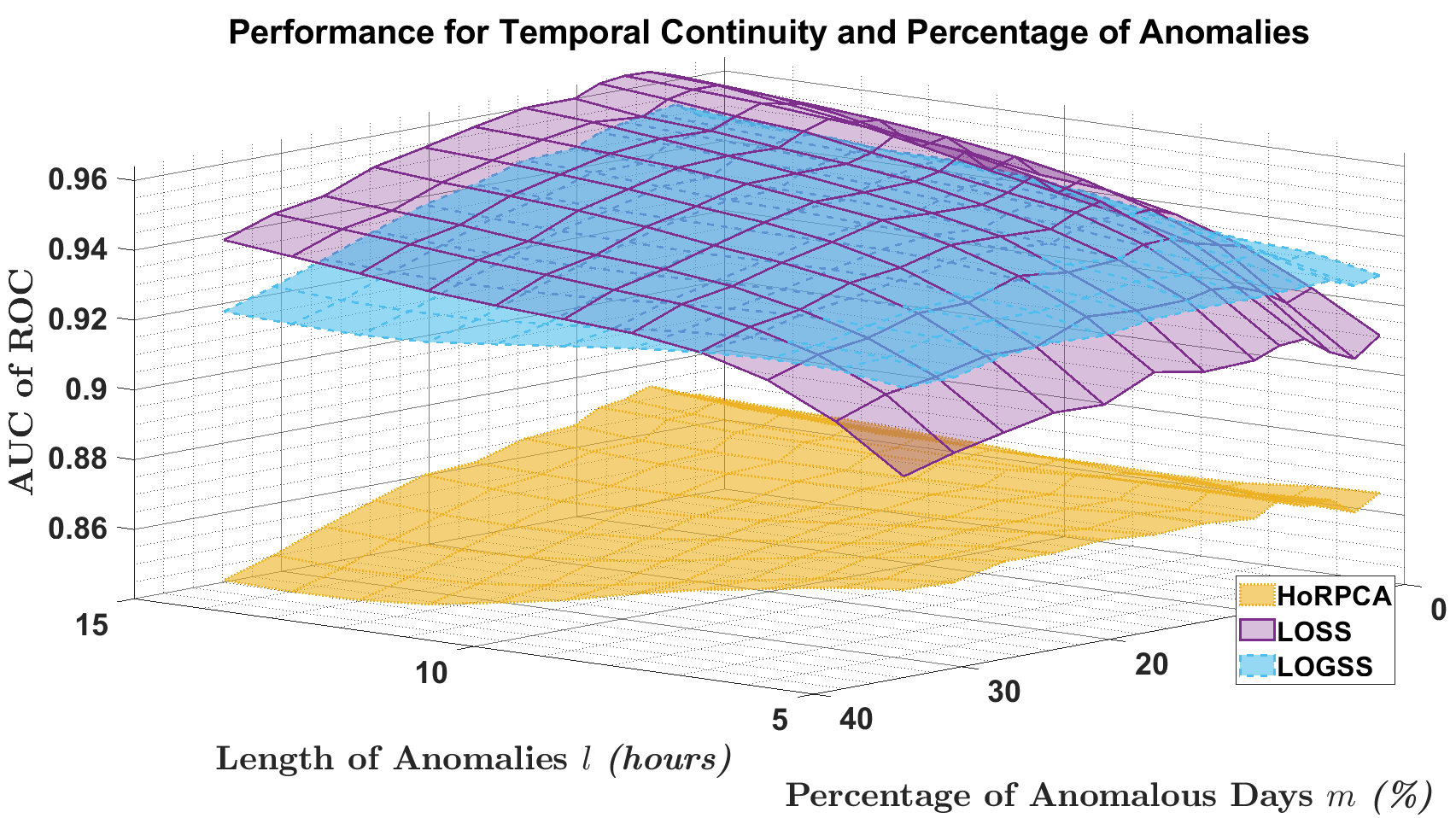}
    \caption{AUC of ROC w.r.t. $l$ and $m$ with $c=2.5$, $P=0\%$.}
    \label{fig:len_num_anom}
    \vspace{-.5em}
\end{figure}

\vspace{-.5em}
First, we evaluated the effect of the length $l$ and the percentage $m$, \textit{i.e.} denseness, of anomalies in synthesized data. For these experiments, we set $c=2.5$ and $P=0\%$. From Table \ref{tab:nyc_auc} and Fig. \ref{fig:len_num_anom}, it can be seen that as $l$ increases, the performance of LOGSS and LOSS improves while HoRPCA's performance does not show a significant change.  This is due to the fact that the temporal total variation regularization will become more suited to the observed data as the anomalies become more temporally persistent, \textit{i.e.} when $l$ increases. Although LOSS performs slightly better than LOGSS when $l$ is large, for low $l$, it drastically underperforms which is not the case for LOGSS. With increasing $m$, all methods perform worse due to the assumption of sparsity for the anomalies. 

In Table \ref{tab:nyc_auc}, we also report the mean AUC values for parameters $c, P$ for all methods, to illustrate the effect of anomaly amplitude and missing data. The proposed method and LOSS have higher anomaly detection accuracy compared to EE and HoRPCA which illustrates the benefit of tailoring the optimization problem to anomaly structure. In fact, HoRPCA does not perform better than EE in most cases which means that extracting anomalies using HoRPCA does not have a significant improvement compared to using the original data. In particular, LOGSS and LOSS are more sensitive to anomalies as they can detect them with higher accuracy even for low $c$ values.  Since both LOSS and LOGSS incorporate tensor completion, they are also robust against missing data. Finally, LOGSS is up to $7$ times faster than LOSS in all experiments.
\vspace{-1em}
\subsection{Experiments on Real Data}
\vspace{-.5em}
For real data, we selected 20 events in NYC during 2018 such as concerts, national holidays and marathons which would result in significant changes in traffic pattern. After computing the anomaly scores for the different methods, top-$K$ percentage of the tensor elements with the highest anomaly scores are selected.  The events that correspond to the selected tensor elements are classified as detected. In previous work, similar case studies were presented for experiments on real data \cite{zhang2020urban,chen2016fine,zhang2018detecting,zhang2019decomposition}. The results are reported in Table \ref{tab:nyc_real}. It can be seen that LOSS and LOGSS improve the performance of anomaly detection with LOSS providing the best results. The performances of HoRPCA and LOGSS are similar although the latter is faster and better at detecting anomalies earlier. It is important to note that 
event selection is done manually and the selected events may not correspond to the most significant anomalies. Thus, although it is a widely utilized tool in analyzing the performance on real data, the case study approach might not reflect the true performance of the anomaly detection method as effectively as synthetic data.

\begin{table}[]
    \scriptsize{}
    \centering
    \caption{Results for $2018$ NYC Yellow Taxi Data. Columns indicate the percentage of selected points with top anomaly scores. The table entries correspond to the number of events detected at the corresponding percentage.}
    \begin{tabular}{l c c c c c c |l}
    \toprule
         \% &  0.14 & 0.3 & 0.7 & 1 & 2 & 3 & Time(sec)\\
         \midrule
         EE  &  1 & 3 & 9 & 9 & 16 & 18 & 15.6\\ 
         HoRPCA &  0 & 1 & 8 & 15 & 18 & 18& 8.6\\
         LOSS &  13 & 16& 17 & 18 & 20 & 20& 17.2\\ 
         LOGSS &  1 & 6 & 10 & 12 & 18 & 18& 3.7\\
    \bottomrule
    \end{tabular}
    \label{tab:nyc_real}
\end{table}

\vspace{-1em}
\section{Conclusion}
\vspace{-.5em}
In this paper, we proposed a computationally efficient tensor decomposition based anomaly detection method for urban traffic data. The proposed method utilizes a robust tensor decomposition with a temporally smooth sparse part to better model the anomaly structure. Low-rank tensor recovery is implemented through minimizing graph total variation on similarity graphs constructed across each mode. This approximation circumvents the need for computing a computationally expensive nuclear norm minimization. The resulting optimization problem is solved using ADMM. The proposed method is compared to EE, HoRPCA and a recently introduced method for anomaly detection with nuclear norm minimization, \textit{i.e.} LOSS.

For both synthetic and real data, LOGSS outperforms other methods in terms of computational efficiency. Experiments on synthetic data reveal that when anomalies are longer in duration, the proposed method and LOSS, which also utilizes a temporal smoothness regularization, perform better. LOGSS also outperforms LOSS when anomalies are shorter in duration. Although LOSS performs better in real data, LOGSS shows similar performance with shorter run time.  

\normalsize

\bibliographystyle{IEEEtran}
\bibliography{main}

\end{document}